%% file: main.tex
\documentclass[a4paper]{article}
\usepackage{authblk}  
\usepackage{cite}
\usepackage{amsmath,amssymb,amsfonts}
\usepackage{algorithmic}
\usepackage{url}
\usepackage{graphicx}
\usepackage{textcomp}

\usepackage{multirow,array}
\usepackage{color}  

\newenvironment{red-color}{\par\color{red}}{\par}

\DeclareGraphicsExtensions{.eps,.pdf}

\usepackage{textcomp,marvosym}

\usepackage{nameref,hyperref}

\usepackage{booktabs}
\usepackage{bm}
\usepackage{ulem}
\usepackage{rotating}

\begin{document}

\title{\Large {\bf A heuristic to determine the initial gravitational constant of the GSA}}

\author{\normalsize
Alfredo J.\ P.\ Barbosa\textsuperscript{1}, 
Edmilson M.\ Moreira\textsuperscript{2*},
Carlos H.\ V.\ Moraes\textsuperscript{2}, \\
Ot\'{a}vio A.\ S.\ Carpinteiro\textsuperscript{2}}

\affil{\normalsize 
\textbf{1} Santa Luzia Faculty, R.\ 21 de Abril, SN, Santa In\^{e}s, MA, 65300-106, Brazil \\
\textbf{2} Research Group on Systems and Computer Engineering, Federal University of Itajub\'{a}, Av.\ BPS 1303, Itajub\'{a}, MG, 37500--903, Brazil \\
\small \it \textbf{*} Corresponding author: Edmilson M.\ Moreira (e-mail: edmarmo@unifei.edu.br)
}

\date{April, 2022}
\maketitle
\thispagestyle{empty}

\section*{Abstract}

\input{abst.txt}

\input{secs.txt}


\section*{Acknowledgments}
This research was supported by CAPES, Brazil.

\bibliographystyle{plain}
\bibliography{References}

\end{document}

%% file: abst.txt
The \textit{Gravitational Search Algorithm (GSA)} is an optimization algorithm based on Newton's laws of gravity and dynamics. Introduced in 2009, the GSA already has several versions and applications. However, its performance depends on the values of its parameters, which are determined empirically. Hence, its generality is compromised, because the parameters that are suitable for a particular application are not necessarily suitable for another. This paper proposes the Gravitational Search Algorithm with Normalized Gravitational Constant (GSA-NGC), which defines a new heuristic to determine the initial gravitational constant of the GSA. The new heuristic is grounded in the Brans-Dicke theory of gravitation and takes into consideration the multiple dimensions of the search space of the application. It aims to improve the final solution and reduce the number of iterations and premature convergences of the GSA. The GSA-NGC is validated experimentally, proving to be suitable for various applications and improving significantly the generality, performance, and efficiency of the GSA.

%% file: secs.txt
\section{Introduction}
\label{sec:intro}

The \textit{Gravitational Search Algorithm (GSA)} \cite{rashedi} is an optimization algorithm based on Newton's laws of gravity and dynamics. Introduced in 2009, the GSA already has several versions, such as the Binary GSA (BGSA) \cite{rashedi1}, Multi-Objective GSA (MO-GSA) \cite{hassanzadeh}, Improved GSA (IGSA) \cite{sarafrazi}, Fuzzy GSA (FGSA) \cite{saeidi}, Discrete GSA (DGSA) \cite{dowlatshahi}, Black Hole GSA (BH-GSA) \cite{doraghinejad,doraghinejad1}, Niche GSA (NGSA) \cite{yazdani}, GSA with Negative Mass (GSAN) \cite{khajooei} and the Fitness Varying Gravitational Constant GSA (FVGGSA) \cite{bansal}, among others. It likewise has several applications in several areas, such as in electric power systems \cite{duman,duman1,shunli,shunli1,chen,packiasudha,beigvand,yashemi,marzbanda,moeini}, modeling of digital filters \cite{rashedi2}, and training of neural networks \cite{mirjalili}, among others.

However, the performance of the GSA depends on the values of its parameters --- number of particles, initial value and rate of change of the gravity, etc. --- that are determined empirically \cite{sabri}. Regarding the initial gravitational constant, there is no formula to determine its best value for each application. In fact, in all of the versions of the GSA and in all of the applications mentioned, the constant used is determined empirically, through a series of trials and corrections. Therefore, the generality of the algorithm is compromised, because a value that is suitable for a specific application is not necessarily suitable for another.

This paper proposes the \textit{Gravitational Search Algorithm with Normalized Gravitational Constant (GSA-NGC)}, which defines a new heuristic to determine the initial gravitational constant of the GSA. The new heuristic is grounded in the Brans-Dicke theory of gravitation \cite{brans} and takes into consideration the multiple dimensions of the search space of the application. It aims to improve the final solution and reduce the number of iterations and premature convergences of the GSA. The GSA-NGC is validated experimentally, proving to be suitable for various applications and improving the generality, performance, and efficiency of the GSA.

The remainder of the paper is organized as follows. The second section describes the GSA. The third section provides an overview of the more recent versions of GSA. The fourth section introduces the proposed heuristic. The fifth section describes the experiments and discusses the results. The last section presents the main conclusions of the paper, and indicates some directions for future work.

\section{GSA}
\label{sec:gsa}

The GSA is an evolutionary optimization algorithm in which each solution is represented by one particle, according to \autoref{eq:particle}

\begin{equation}
x_i(t)=(x_{i1}(t), x_{i2}(t), ..., x_{iD}(t)), \quad i=1,\ ...,\ P
\label{eq:particle}
\end{equation}

\noindent where $x_i(t)$ is the position of the particle $i$ in the iteration $t$, $x_{ij}(t)$ is the dimension $j$ of the position of the particle $i$ in iteration $t$, $D$ is the number of dimensions of the search space of the application, and $P$ is the number of particles of the algorithm.

The initial position of each particle is randomly determined in the search space, according to \autoref{eq:initialization}

\begin{equation}
x_i(0)=u(D)\circ(L^+-L^-)+L^-
\label{eq:initialization}
\end{equation}

\noindent where $u(D)$ is a vector of dimension $D$ of random variables in the interval $[0,\ 1]$, and $L^+$ and $L^-$ are vectors of dimension $D$. $L_j^+$ and $L_j^-$ are respectively the maximum and minimum limits of the dimension $j$ of the search space. $D$ is the number of dimensions of the search space of the application.

The particles attract each other according to the Newton's law of gravity, and move according to the Newton's second law. The fitness of a solution (i.e., a particle) is given by its mass, which is updated in each iteration. The time unit of the simulation is the iteration. Each iteration of the GSA comprises the following steps:

\begin{enumerate}

\item The mass of each particle is calculated in three phases:

\begin{enumerate}

\item The value of the objective function $o_i$ for each particle $i$ is calculated according to \autoref{eq:objective}

\begin{equation}
o_i(t)=f(x_i(t))
\label{eq:objective}
\end{equation}

\noindent where $f(x_i(t))$ is the value of the objective function of the particle $x_i$ in the iteration $t$.

\item The fitness of each solution is calculated according to \autoref{eq:fitness}

\begin{equation}
q_i(t)=\frac{o_i(t)-w(t)}{b(t)-w(t)}
\label{eq:fitness}
\end{equation}

\noindent where $q_i(t)$ is the fitness of the solution $i$ in the iteration $t$, $o_i(t)$ is the value of the objective function of the solution $i$ in the iteration $t$, and $b(t)$ and $w(t)$ are respectively the best and worst values of the objective function in the iteration $t$.

Therefore, the dynamics of GSA depends neither on the fact that the objective function is either maximized or minimized nor on its maximum and minimum values.

\item There are three types of masses --- active gravitational mass, passive gravitational mass, and inertial mass. However, GSA makes use only of the active gravitational mass, which will be, from now on, referred to as simply mass. The other two masses --- passive gravitational and inertial --- have values equal to 1, and thus do not appear in the equations of GSA. The mass of each particle is calculated according to \autoref{eq:mass}

\begin{equation}
m_i(t)=\frac{q_i(t)}{\sum_{j=1}^P{q_j(t)}}
\label{eq:mass}
\end{equation}

\noindent where $m_i(t)$ is the mass of the particle $i$ in the iteration $t$, $q_i(t)$ is the fitness of the solution $i$ in the iteration $t$, and $P$ is the number of particles of the algorithm.

Therefore, the dynamics of GSA does not depend on the number of particles as well, because the sum of the masses of all particles is always equal to 1.

\end{enumerate}

\item The total force on each particle is also calculated in three phases:

\begin{enumerate}

\item The gravity\footnote{In addition to gravity, the term \textit{gravitational constant} is also used. However, \textit{gravitational constant} is an erroneous term, for the gravity is not in fact a constant, but a decreasing function in GSA.} is proportional to the initial gravitational constant according to \autoref{eq:variation}

\begin{equation}
G(t)=G_0e^{-\alpha{t}/T}
\label{eq:variation}
\end{equation}

\noindent where $G(t)$ is the gravity in the iteration $t$, $G_0$ is the initial gravitational constant, $\alpha$ is the constant which determines the magnitude of the variation of the gravity, and $T$ is the maximum number of iterations of the algorithm.

\item The force of the particle $j$ on the particle $i$ is calculated according to \autoref{eq:force}

\begin{equation}
F_{ij}(t)=G(t)\frac{(x_j(t)-x_i(t))}{|x_j(t)-x_i(t)|}m_j(t)
\label{eq:force}
\end{equation}

\noindent where $F_{ij}(t)$ is the force of the particle $j$ on the particle $i$ in the iteration $t$, $x_i(t)$ is the position of the particle $i$ in the iteration $t$, $m_j(t)$ is the mass of the particle $j$ in the iteration $t$, and $G(t)$ is the gravity in the iteration $t$.

Thus, the force of the particle $j$ on the particle $i$ does not depend on the distance between both particles, but on the direction between them. The \autoref{eq:force} is then simplified, resulting in \autoref{eq:force1}

\begin{equation}
F_{ij}(t)=G(t)<x_j(t)-x_i(t)>m_j(t)
\label{eq:force1}
\end{equation}

\noindent where $<x_j(t)-x_i(t)>$ is an unit vector in the direction between particles $j$ and $i$.

\item The total force on each particle $i$ is calculated according to \autoref{eq:total}

\begin{equation}
\label{eq:total}
F_i(t)=\sum_{j=1}^P(u(D)\circ{F_{ij}(t)}), \quad j\neq{i}
\end{equation}

\noindent where $F_i(t)$ is the total force on the particle $i$ in the iteration $t$, $F_{ij}(t)$ is the force of the particle $j$ on the particle $i$ in the iteration $t$, $u(D)$ is a vector of dimension $D$ of random variables in the interval $[0,\ 1]$, $P$ is the number of candidate solutions of the algorithm, and $D$ is the dimension of the search space of the application.

Combining \autoref{eq:force1} with \autoref{eq:total} produces \autoref{eq:total1}

\begin{equation}
F_i(t)=G(t)\sum_{j=1}^P(u(D)\circ<x_j(t)-x_i(t)>m_j(t)), \quad j\neq{i}
\label{eq:total1}
\end{equation}

\noindent in which it is possible to observe that the total force on particle $i$ is proportional to the gravity.

\end{enumerate}

\item Finally, the position of each particle is updated in two phases:

\begin{enumerate}

\item Since the inertial mass of each particle is equal to 1 and the acceleration of each particle is equal to the total force on it, the velocity $v_i(t)$ of each particle $i$, in the iteration $t$, is updated according to \autoref{eq:velocity}

\begin{equation}
v_i(t)=u(D)\circ{v_i(t-1)+F_i(t)}
\label{eq:velocity}
\end{equation}

\noindent where $F_i(t)$ is the total force on the particle $i$ in the iteration $t$, $u(D)$ is a vector of dimension $D$ of random variables in the interval $[0,\ 1]$, and $D$ is the dimension of the search space of the application.

\item The position of each particle is updated according to \autoref{eq:position}

\begin{equation}
x_i(t)=x_i(t-1)+v(t)
\label{eq:position}
\end{equation}

Whenever a particle leaves the search space, it is placed again in the search space either in a random position or in a limit of the search space.

The execution of the algorithm is ended when the stop condition --- number of iterations, time limit, or any other --- is reached. The best solution is then returned.

\end{enumerate}

\end{enumerate}

The main characteristics of global optimization algorithms, and in particular of the GSA, are the local search (\textit{exploitation}) and the global search (\textit{exploration}). The exploration and exploitation of the GSA are determined by the movement of the particles. When the velocity of a particle is high, independently of its direction, it performs exploration. When its velocity is low, it performs exploitation. The acceleration of any particle is proportional to the gravity, and consequently, exploration is increased by a large gravity value and exploitation by a small one. Moreover, the gravity is a decreasing function. Thus, exploration is high in the first iterations and low in the last ones, and exploitation is low in the first iterations and high in the last ones.

GSA makes use of an elitist strategy. In this strategy, only some particles --- called active particles --- are taken into consideration in \autoref{eq:total}, yielding \autoref{eq:elitist}

\begin{equation}
F_i(t)=\sum_{j=1}^{K(t)}(u(D)\circ{F_{i, best(j)}(t)}), \quad best(j)\neq{i}
\label{eq:elitist}
\end{equation}

\noindent where $F_i(t)$ is the total force on the particle $i$ in the iteration $t$, $F_{i, best(j)}(t)$ is the force of the particle $best(j)$ on the particle $i$ in the iteration $t$, $best(j)$ is the $j$-th best solution, $u(D)$ is a vector of dimension $D$ of random variables in the interval $[0,\ 1]$, $K(t)$ is the number of active particles in the iteration $t$, and $D$ is the dimension of the search space of the application.

The number of active particles $K(t)$ should be maximum (i.e., equal to the total number of particles of the algorithm) in the first iteration, and should decrease, step-by-step through the iterations, to the value equal to 1 in the last iteration, according to \autoref{eq:elitist1}

\begin{equation}
K(t)=P-\lfloor(P-1)(t/T)\rceil
\label{eq:elitist1}
\end{equation}

Thus, in the first iterations, when many particles are active, many solutions are explored, increasing the exploration of the GSA. In the last iterations, however, when few particles are active, few solutions are explored, increasing its exploitation.

To sum up, a solution is represented by a particle. The particles attract each other and move in the search space. Their positions and masses are updated in each iteration. The best solution obtained is returned when a stop condition is reached.

\section{Related Work}
\label{sec:version}

The works most closely related to GSA-NGC are all other versions of the GSA available in the literature. These works introduce changes in the GSA in order to improve its performance or other characteristics.

This section discusses the more recent versions of GSA. They are the Binary GSA (BGSA) \cite{rashedi1}, Multi-Objective GSA (MO-GSA) \cite{hassanzadeh}, Improved GSA (IGSA) \cite{sarafrazi}, Fuzzy GSA (FGSA) \cite{saeidi}, Discrete GSA (DGSA) \cite{dowlatshahi}, Black Hole GSA (BH-GSA) \cite{doraghinejad,doraghinejad1}, Niche GSA (NGSA) \cite{yazdani}, GSA with Negative Mass (GSAN) \cite{khajooei} and Fitness Varying Gravitational Constant GSA (FVGGSA) \cite{bansal}.

The proposed GSA-NGC is different from all these versions because it is the first (and unique) version of GSA that proposes an heuristic (NGC) to determine the initial gravitational constant of the GSA. In addition, NGC may be combined with the existent versions of GSA in order to improve their performance and robustness.

\subsection{Multi-objective GSA}

Hassanzadeh and Rouhani \cite{hassanzadeh} proposed a version of GSA adapted to multi-objective applications --- \textit{Multi-Objective GSA (MO-GSA)} --- in 2010. MO-GSA uses some strategies of multi-objective optimization, such as the uniform mutation operation, and the elitist strategy. The authors make use of two metrics of the multi-objective optimization --- spacing and generational distance.

The spacing $S$ measures the degree of uniformity of the distance between the solutions, according to \autoref{eq:mogsa}

\begin{equation}
S=\sqrt{\frac{1}{N-1}\sum_{i=1}^N(d_i-\overline{d})^2}, \quad i=1,\ 2,\ ...,\ N
\label{eq:mogsa}
\end{equation}

\noindent where $d_i$ is given by \autoref{eq:mogsa1}

\begin{equation}
d_i=min\sum_{k=1}^O|f_k(x_i)-f_k(x_j)|, \quad j=1,\ 2,\ ...,\ N,\ j\neq{i}
\label{eq:mogsa1}
\end{equation}

\noindent where $f_k(x_i)$ is the value of the objective function $k$ of the particle $i$, $O$ is the number of objective functions of the application, $\overline{d}$ is the mean of all variables $d_i$, and $N$ is the number of particles.

The generational distance $GD$ measures the distance between the solutions obtained and the Pareto frontier, according to \autoref{eq:mogsa2}

\begin{equation}
GD=\frac{1}{N}\sqrt{\sum_{i=1}^N{p_i^2}}
\label{eq:mogsa2}
\end{equation}

\noindent where $p_i$ is the distance between the solution $i$ and the Pareto frontier.

The MO-GSA was compared to the simple multi-objective PSO (SMOPSO) and to the Pareto archived evolution strategy (PAES) in experiments with three reference functions. The results achieved by MO-GSA were better than those achieved by the two other algorithms. However, the computational cost of the MO-GSA is higher.

\subsection{Improved GSA}

Sarafrazi et al.\ \cite{sarafrazi} proposed in 2011 an improved version of GSA --- \textit{Improved GSA (IGSA)} ---, which uses a disruption operation. The disruption consists in breaking apart the particles of a group of particles united by gravity, whenever the group moves closer to a particle with a large mass. Thus, when the particle $i$ of the group moves closer to the ``bulky'' particle $j$, the disruption operation is given by \autoref{eq:igsa}

\begin{equation}
\frac{R_{ij}}{R_i^*}<C
\label{eq:igsa}
\end{equation}

\noindent where $R_{ij}$ is the distance between the particle $i$ and the particle $j$, $R_i^*$ is the distance between the particle $i$ and the best solution of the iteration, and $C$ is the variable employed to control the exploration and exploitation of the disruption operation.

When a particle leaves the group, its position is updated, according to \autoref{eq:igsa1}

\begin{equation}
x_i=
\begin{cases}
random(-0.5,\ +0.5)R_{ij}x_i' \quad & if\ R_i^*\geq1\\
{random(-0.5,\ +0.5)}\rho x_i'+x_i' \quad & otherwise
\end{cases}
\label{eq:igsa1}
\end{equation}

\noindent where $random(min, max)$ is a random variable in the interval $[min, max]$, $x_i'$ is the position of the particle $i$ before the disruption, and $x_i$ is the position of the particle $i$ after the disruption. The purpose of the variable $\rho$ is not described by the authors.

The IGSA was compared with the GSA, with the real genetic algorithm (RGA), and with the particle swarm optimization (PSO) in experiments with thirteen well-known reference functions. Its results were better than those from the other algorithms on most experiments.

\subsection{Fuzzy GSA}

Saeidi-Khabisi and Rashedi \cite{saeidi} proposed the \textit{Fuzzy Gravitational Search Algorithm (FGSA)} in 2012. In the FGSA, the fuzzy logic is employed to control the constant $\alpha$ of \autoref{eq:variation}, and to control the exploration and exploitation of the GSA through two metrics --- population diversification and convergence measurement.

The population diversification $PD$ is given by \autoref{eq:fgsa}

\begin{equation}
PD=\frac{R_{ave}-R_{min}}{R_{max}-R_{min}}
\label{eq:fgsa}
\end{equation}

\noindent where $R_{max}$ and $R_{min}$ are respectively the best and worst solution, $R_i^*$ is the distance between the solution $i$ and $R_{max}$, $R_{ave}$ is the mean of all distances $R_i^*$.

The convergence measurement $CM$ of the solutions, from a iteration to the next, is given by \autoref{eq:fgsa1}

\begin{equation}
CM=\frac{fit_{ave}(t-1)-fit_{ave}(t)}{fit_{ave}(t)}
\label{eq:fgsa1}
\end{equation}

\noindent where $fit_{ave}(t)$ is the mean of the fitnesses of the solutions in the iteration $t$.

The FGSA was compared to the SGSA in experiments with some well-known reference functions. Its results were a little better than those from SGSA. However, the FGSA has a higher computational cost.

\subsection{Discrete GSA}

Dowlatshahi et al.\ \cite{dowlatshahi} also proposed another version of the GSA --- \textit{Discrete GSA (DGSA)} ---, suitable for discrete applications, in 2014. In the DGSA, the movement of each particle, defined in GSA as velocity (\autoref{eq:velocity}), was divided into two diferent types of movement --- one independent and another dependent of the position of the particles. The independent movement length (IML) is given by $u(D)\circ{v_i(t-1)}$. In its turn, the dependent movement length (DML) is given by $F_i(t)$. Each movement was substituted by a discrete search operator.

The DGSA was compared to other algorithms in experiments with the travelling salesman application. Its results indicate that it may be used in discrete applications.

\subsection{Black hole GSA 1}

Doraghinejad et al.\ \cite{doraghinejad} proposed a version of GSA --- \textit{Black hole GSA 1 (BH-GSA1)} --- in 2012, which makes use of a black-hole operation. According to the relativity theory, the gravitational force inside a black hole is huge.

In the BH-GSA1, heavier particles become black holes. The radius $R_s$ of a black hole is given by \autoref{eq:bhgsa}

\begin{equation}
R_s=M*log(t)
\label{eq:bhgsa}
\end{equation}

\noindent where $M$ is the mass of the black hole.

The other particles are divided into two categories --- heavy particles and light particles. Whenever a heavy particle lies inside the radius of a black hole, its position is updated in accordance with \autoref{eq:bhgsa1}

\begin{equation}
x_i(t)=x_i(t)\left(exp\left(-\frac{t-1}{2M}\right)+1\right)
\label{eq:bhgsa1}
\end{equation}

Whenever a light particle lies inside the radius of a black hole, its position is updated in accordance with \autoref{eq:bhgsa2}

\begin{equation}
x_i(t)=x_i(t)R\left(\frac{1}{log(t-1)+1}\right)
\label{eq:bhgsa2}
\end{equation}

\noindent where $R$ is the distance between the particle and the black hole.

The BH-GSA1 was compared with the SGSA and with the IGSA in experiments with seven well-known unimodal reference functions. The results indicated that the BH-GSA1 is better than the other two algorithms in unimodal applications.

\subsection{Black hole GSA 2}

Doraghinejad and Nezamabadi-pour \cite{doraghinejad1} proposed another version of GSA --- \textit{Black hole GSA 2 (BH-GSA2)} --- in 2014, which also makes use of the black-hole operation.

In the BH-GSA2, only the heaviest particle becomes a black hole. The black hole has two radius --- $R_s$, given by \autoref{eq:bhgsa}, and $R_s'$, given by \autoref{eq:bhgsa3}

\begin{equation}
R_s'=GMv^2/t
\label{eq:bhgsa3}
\end{equation}

\noindent where $v$ is the velocity of the black hole.

The other particles are also divided into heavy particles and light particles. Whenever a heavy particle lies inside the radius $R_s$ of a black hole, its position is updated in accordance with \autoref{eq:bhgsa4}

\begin{equation}
x_i^d(t)=x_i^d(t)+random(x_{BH}(t-1)-x_i(t-1))
\label{eq:bhgsa4}
\end{equation}

\noindent where $x_{BH}$ is the position of the black hole.

Whenever a light particle lies inside the radius $R_s'$ of a black hole, its position is updated in accordance with \autoref{eq:bhgsa5}

\begin{equation}
x_i^d(t)=random(x_i^d(t-1)(r_i/R_s'))
\label{eq:bhgsa5}
\end{equation}

\noindent where $r_i$ is the distance between the particle $i$ and the black hole.

The BH-GSA2 was compared with four algorithms --- RGA, PSO, SGSA, and IGSA --- in experiments with several reference functions. The results indicated that the convergence of the BH-GSA2 is similar to that of IGSA and better than that of the other algorithms, in general.

\subsection{Niche GSA}

Yazdani et al.\ \cite{yazdani} proposed a version of the GSA --- \textit{Niche GSA (NGSA)} --- suitable for multimodal applications, in 2014. In the NGSA, each particle has a niche which comprises $K$ particles. The particle $i$ belongs to the niche of the particle $j$ only if $i$ is one of the $K$ particles closer to $j$. The gravitational mass of each particle is a vector, according to \autoref{eq:ngsa}

\begin{equation}
m_i(t)=(m_{i1}(t),\ m_{i2}(t),\ ...,\ m_{iN}(t))
\label{eq:ngsa}
\end{equation}

\noindent where $m_{ij}(t)$ is the mass of the particle $i$ in the niche of the particle $j$ in the iteration $t$. $m_{ij}(t)$ is given by \autoref{eq:ngsa1}

\begin{equation}
m_{ij}(t)=
\begin{cases}
\frac{o_i(t)-w_j(t)}{b_j(t)-w_j(t)} \quad & \text{if $i$ belongs to niche of $j$}\\
0 \quad & \text{otherwise}
\end{cases}
\label{eq:ngsa1}
\end{equation}

\noindent where $o_i(t)$ is the value of the objective function for the solution (i.e., particle) $i$, $b_j(t)$ is the value of the objective function for the best solution in the niche of the particle $j$, and $w_j(t)$ is the value of the objective function for the worst solution in the niche of $j$.

The number of particles $K(t)$ inside each niche varies through the iterations, according to \autoref{eq:ngsa2}

\begin{equation}
K(t)=round([K_i-(K_i-K_f)t/T]N)
\label{eq:ngsa2}
\end{equation}

\noindent where $K_i$ and $K_f$ are respectively the initial and the final number of particles inside the niches.

In the NGSA, the position of a particle is updated only if the solution of the new position is better than the former.

The NGSA was compared with some algorithms --- deterministic crowding, sequential niche, and some versions of PSO ---, suitable for multimodal applications, in experiments with several multimodal reference functions. The results indicated that the NGSA is suitable for multimodal applications, but is sensitive to the values assigned to $K_i$ and $K_f$.

\subsection{GSA with negative masses}

Khajooei and Rashedi \cite{khajooei} proposed a version of GSA --- \textit{GSA with Negative Masses (GSAN)} --- in 2016, which takes into consideration the anti-gravity. One of the shortcomings of the GSA is premature convergence, which arises from the fact that, owing to the gravitational force, each particle only attracts each other, but never repels each other. The GSAN is based on the supposition that the gravitational force has an inverse force, called anti-gravity. Thus, in the GSAN, each particle with positive mass repels each other with negative mass, and vice-versa.

The mass of a particle in GSAN is given by \autoref{eq:gsan}

\begin{equation}
m_i(t)=\frac{q_i(t)}{\sum_{j=1}^N|q_j(t)|}
\label{eq:gsan}
\end{equation}

\noindent where $q_i(t)$ is the fitness of the solution $i$, according to \autoref{eq:gsan1}

\begin{equation}
q_i(t)=\frac{o_i(t)-w(t)}{b(t)-w(t)}-C(t)
\label{eq:gsan1}
\end{equation}

Therefore, if the fitness of the solution $i$ in the iteration $t$ is larger than the value of $C$ in the iteration $t$, then the mass of the particle $i$ is positive, otherwise it is negative. $C$ is given by \autoref{eq:gsan2}

\begin{equation}
C(t)=C_0e^{-\beta{t}/T}
\label{eq:gsan2}
\end{equation}

The GSAN was compared to the SGSA in experiments with some reference functions, but its results were not significantly better than those from the SGSA.

\subsection{Fitness Varying Gravitational Constant GSA}

Bansal et al.\ \cite{bansal} proposed a recent version of GSA --- \textit{Fitness Varying Gravitational Constant GSA (FVGGSA)} --- in 2018. The execution of FVGGSA comprises several phases. In each phase $p$, the gravity $G'$ is given by \autoref{eq:fvggsa}

\begin{equation}
G'(t)=Ze^{-\alpha{t/\eta}}
\label{eq:fvggsa}
\end{equation}

\noindent where $G'(t)$ is the gravity in the iteration $t$ of phase $p$, $\eta$ is the maximum number of iterations of phase $p$ and $Z$ is the scale constant of phase $p$. The constant $Z$ is calculated by \autoref{eq:fvggsa1}

\begin{equation}
Z(x)=1408e^{-0.00529x}
\label{eq:fvggsa1}
\end{equation}

\noindent where $x$ is the mean iteration of phase $p$.

A \textit{fitness varying gravitational constant (i.e., fitness varying gravity)} is calculated for each particle, according to \autoref{eq:fvggsa2}

\begin{equation}
G_i(t)=G'(t)(1.45-prob_i)
\label{eq:fvggsa2}
\end{equation}

\noindent where the variable $prob_i$ is given by \autoref{eq:fvggsa3}

\begin{equation}
prob_i=\frac{0.9*fit(i)}{max\ fit}+0.1
\label{eq:fvggsa3}
\end{equation} 

\noindent where $fit(i)$ is the quality of the solution $i$, according to \autoref{eq:fvggsa4}

\begin{equation}
fit(i)=\begin{cases}
1-f_i, \quad & f_i<0 \\
\frac{1}{1+f_i}, \quad & f_i\geq0
\end{cases}
\label{eq:fvggsa4}
\end{equation}

\noindent where $f_i$ is the value of the objective function of the solution $i$.

The FVGGSA was compared to four algorithms --- SGSA, Chaotic GSA (CGSA), Biogeography-based Optimization (BBO) and Disruption BBO (DBBO) --- on two sets of reference functions. The FVGGSA achieved better results than the other four algorithms.

\section{Proposed heuristic}
\label{sec:heuristic}

As mentioned before (\autoref{sec:intro}), the performance of the \textit{GSA} depends on the initial gravitational constant, whose value is determined empirically. In this section, the heuristic \textit{Normalized Gravitational Constant (NGC)} is proposed to determine the value of the initial gravitational constant.

The section is divided into three subsections. The first (\autoref{ssec:brans}) presents the Brans-Dicke theory. The second (\autoref{ssec:motion}) evaluates the motion of particles in the GSA. Lastly, the third (\autoref{ssec:ngc}) describes the heuristic \textit{NGC} and determines the value of its proportionality constant $\beta$.

\subsection{Brans-Dicke theory}
\label{ssec:brans}

In his theory, Isaac Newton proposed two types of space --- absolute and relative ---, and two types of motion --- absolute and relative. The absolute space is an immutable, immovable, and matter-independent physical structure. The relative space, in its turn, is the space determined by the relative positions of the physical bodies. The absolute and relative motion is the motion of a physical body in relation to the absolute and relative space, respectively \cite[p. 77]{newton}.

Conversely, in his theory, Ernst Mach proposed only one type of space --- relative ---, and only one type of motion --- relative. The relative space (i.e., its geometric characteristics) is determined by the arrangement of matter in the universe. For Mach, the absolute space and the absolute motion are mental constructs which cannot be produced experimentally. All principles of mechanics are given by experimental knowledge about the relative positions and the relative motions of the physical bodies. Therefore, it is not possible to extend these principles beyond the experimental limits \cite[p. 229]{mach}.

In addition to the space geometry, the supporters of the Machian theory conjecture that two forces of dynamics --- inertial force and gravitational force --- are also determined by the arrangement of matter in the universe. The major limitation of this conjecture is that it does not establish a quantitative relation between these forces and the mass of the universe \cite[p. 124]{inverno}. Several theories were proposed to overcome this limitation, among which the Brans-Dicke theory stands out \cite{brans}.

The Brans-Dicke theory of gravitation states that the gravity (i.e., gravitational force) is given by \autoref{eq:machian}

\begin{equation}
G\sim\frac{Rc^2}{M}
\label{eq:machian}
\end{equation}

\noindent where $G$ is the gravity, $R$ and $M$ are the radius and mass of the visible universe, respectively, and $c$ is the speed of light in vacuum.

Therefore, \autoref{eq:machian} states that the gravity is directly proportional to the radius of the visible universe. In other words, it is directly proportional to the dimension of the visible universe, i.e.,

\begin{equation}
G\propto{D}
\label{eq:machian1}
\end{equation}

\noindent where $G$ is the gravity and $D$ is the dimension of the visible universe.

\subsection{Motion of particles in the GSA}
\label{ssec:motion}

Through the equations of GSA, it is possible to verify that the velocity of each particle is directly proportional to the total force on the particle (\autoref{eq:velocity}). The total force on each particle (\autoref{eq:total}) is directly proportional to the gravity (\autoref{eq:force1}), which, in its turn, is directly proportional to the initial gravitational constant (\autoref{eq:variation}). Thus, 

\begin{equation}
{v_i}\propto{F_i}\propto{G}\propto{G_0}
\label{eq:relation}
\end{equation}

\noindent where $v_i$ is the velocity of particle $i$, $F_i$ is the total force on the particle $i$, $G$ is the gravity, and $G_0$ is the initial gravitational constant.

\begin{figure}[htb]
\center
\includegraphics[width=0.75\linewidth]{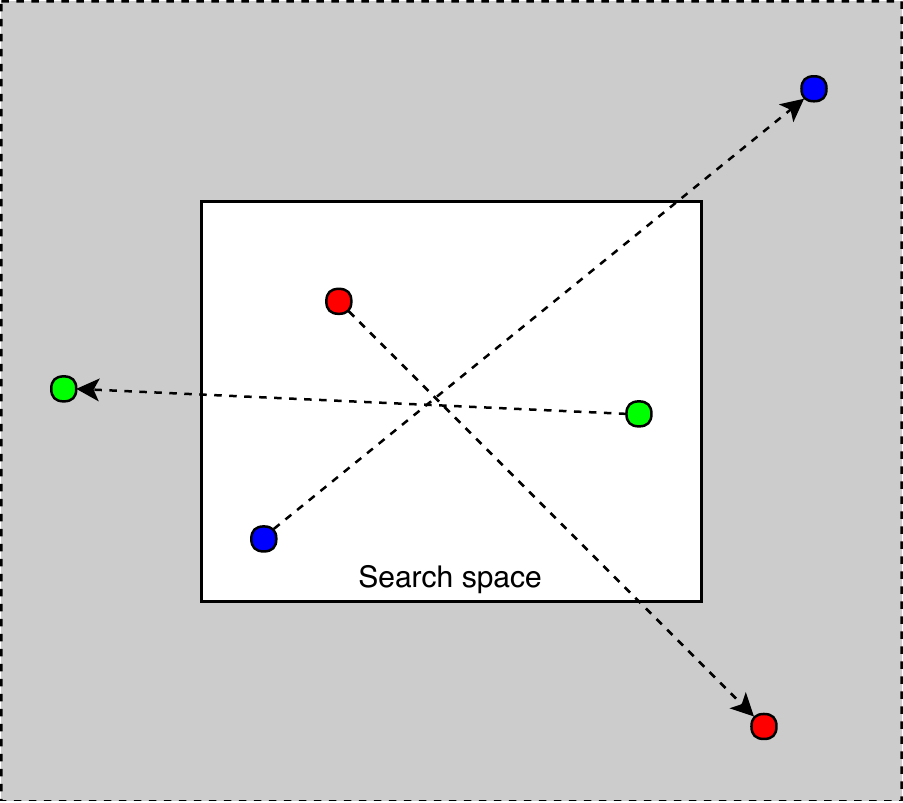}
\caption{Exodus of particles}
\label{fig:condition1}
\end{figure}

Moreover, if the velocity of the particles is high in the first iteration, many particles will leave the search space of the application, as shown in \autoref{fig:condition1}. This exodus of particles undermine the exploration phase of the algorithm, persisting for many iterations, until the gravity decays to an adequate value over the iterations.

Conversely, if the velocity of the particles is low in the first iteration, the exploration of the search space of the application will be small. In addition, the exploration will be reduced with the decay of the gravity over the iterations, provoking premature convergence or even stagnation of the algorithm.

Therefore, the velocity of each particle in GSA should have an inferior and a superior limit. That means, by \autoref{eq:relation}, that the gravity should have an inferior and a superior limit as well. However, the superior limit (i.e., the value of the initial gravitational constant) is the only one relevant, for the inferior limit (i.e., the final value of the gravitational function), according to \autoref{eq:variation}, derives from the superior limit. Thus, 

\begin{equation}
{v_i}\propto{F_i}\propto{G}\propto{G_0}=\text{superior  limit}
\label{eq:limit}
\end{equation}

\noindent where $v_i$ is the velocity of particle $i$, $F_i$ is the total force on the particle $i$, $G$ is the gravity, $G_0$ is the initial gravitational constant and the superior limit.

\subsection{Heuristic NGC}
\label{ssec:ngc}

As mentioned in \autoref{ssec:motion}, the gravity should be limited by a superior limit, and this limit is the initial gravitational constant itself (\autoref{eq:limit}). Conversely, as mentioned in \autoref{ssec:brans}, the gravity is directly proportional to the dimension of the visible universe (\autoref{eq:machian1}).

The heuristic NGC, proposed to determine the initial gravitational constant of GSA, is grounded in Brans-Dicke theory. The visible universe in Brans-Dicke theory corresponds to the search space of the application in GSA. Since the total mass of the particles in GSA is normalized, the initial gravitational constant $G_0$, given by the heuristic NGC, is thus directly proportional to the dimension of the search space of the application, according to \autoref{eq:heuristic}

\begin{equation}
G_0=\beta*\sum_{i=1}^N(L_i^+ - L_i^-)/N
\label{eq:heuristic}
\end{equation}

\noindent where $\beta$ is the proportionality constant of the relation, $L_i^+$ and $L_i^-$ are, respectively, the maximum and minimum limits of dimension $i$ of the search space, and $N$ is the total number of dimensions of the search space (i.e., the number of variables of the application).

Experiments were performed to determine the value of the proportionality constant $\beta$. For that, the GSA-NGC was coded in language Octave \cite{octave}.

The experiments were conducted, with $\beta$ values equal to 0.125, 0.25, 0.5, 1.0, 2.0, 4.0 and 8.0, on thirteen well-known reference functions compiled by Yao et al.\ \cite{xinyao}. The thirteen functions comprises seven unimodal functions --- Sphere, Schwefel 2.22, Schwefel 1.2, Schwefel 2.21, Rosenbrock, Step, Quartic --- and six multimodal functions --- Schwefel 2.26, Rastrigin, Ackley, Griewank, Penalized 1, Penalized 2. The equations of the thirteen functions are presented in \autoref{tab:reference1} and their optimal values, optimal solutions, minimum and maximum limits of the dimensions of the search space are displayed in \autoref{tab:reference2}. Four values of parameters --- number of particles $P=50$, number of iterations $T=1000$, number of dimensions $D=30$, and $\alpha=20$ --- were employed in the experiments. The results are the means of thirty executions of the GSA-NGC with each value of $\beta$. They are shown in \autoref{tab:normalized}.

\begin{table*}[htb]
\centering
\renewcommand{\arraystretch}{2}
\caption{Reference functions --- the first seven are unimodal, and the remaining six are multimodal}
\label{tab:reference1}
\begin{tabular}{c c}\toprule
{\bf Function} & {\bf Equation} \\\midrule
Sphere & $f(X)=\sum_{i=1}^N{X_i^2}$ \\\midrule
Schwefel 2.22 & $f(X)=\sum_{i=1}^N{|X_i|}+\prod_{i=1}^N{|X_i|}$ \\\midrule
Schwefel 1.2 & $f(X)=\sum_{i=1}^N(\sum_{j=1}^i{x_j})^2$ \\\midrule
Schwefel 2.21 & $f(X)=max(|X_1|, |X_2|, ..., |X_N|)$ \\\midrule
Rosenbrock & $f(X)=\sum_{i=2}^N[100(X_i-X_{i-1}^2)^2+(X_{i-1}-1)^2]$ \\\midrule
Step & $f(X)=\sum_{i=1}^N\lfloor{X_i+0.5}\rfloor^2$ \\\midrule
Quartic & $f(X)=\sum_{i=1}^N(iX_i^4)+u(1)$ \\\midrule
Schwefel 2.26 & $f(X)=-\sum_{i=1}^N(X_isin(\sqrt{|X_i|}))$ \\\midrule
Rastrigin & $f(X)=\sum_{i=1}^N(X_i^2-10cos(2X_i\pi)+10)$ \\\midrule
Ackley & $\begin{aligned}
f(X)= & -20exp(-0.2\sqrt{\frac{1}{N}\sum_{i=1}^N{X_i^2}}) \\
 & -exp(\frac{1}{N}\sum_{i=1}^N{cos(2X_i\pi)})+e+20
\end{aligned}$ \\\midrule
Griewank & $f(X)=\frac{1}{4000}\sum_{i=1}^N{X_i^2}-\prod_{i=1}^N{cos(\frac{X_i}{\sqrt{i}})}+1$ \\\midrule
Penalized 1 & $\begin{aligned}
f(X)= & \frac{\pi}{N}\{10sin^2(y_1\pi)+\sum_{i=2}^N[(y_{i-1}-1)^2(10sin^2(y_i\pi)+1)]+ \\
 & (y_N-1)^2\}+\sum_{i=1}^N{u(X_i, 10, 100, 4)}
\end{aligned}$ \\\midrule
Penalized 2 & $\begin{aligned}
f(X)= & 0.1\{sin^2(3X_1\pi)+\sum_{i=2}^N[(X_{i-1}-1)^2(sin^2(3X_i\pi)+1)]+ \\
 & (X_N-1)^2(sin^2(2X_N\pi)+1)\}+\sum_{i=1}^N{u(X_i, 5, 100, 4)}
\end{aligned}$ \\\bottomrule
\end{tabular}
\end{table*}


\begin{table*}[htb]
\centering
\caption{Reference functions --- optimal values, optimal solutions, minimum and maximum limits of the dimensions of the search space}
\label{tab:reference2}
\begin{tabular}{c c c c c}\toprule
{\bf Function} & {\bf Opt. value} & {\bf Opt. solution} & 
{\bf Min. limit} & {\bf Max. limit} \\\midrule
Sphere & 0 & (0,\ 0,\ \ldots,\ 0) & $-$1.0E$+$2 & $+$1.0E$+$2 \\
Schwefel 2.22 & 0 & (0,\ 0,\ \ldots,\ 0) & $-$1.0E$+$1 & $+$1.0E$+$1 \\
Schwefel 1.2 & 0 & (0,\ 0,\ \ldots,\ 0) & $-$1.0E$+$2 & $+$1.0E$+$2 \\
Schwefel 2.21 & 0 & (0,\ 0,\ \ldots,\ 0) & $-$1.0E$+$2 & $+$1.0E$+$2 \\
Rosenbrock & 0 & (1,\ 1,\ \ldots,\ 1) & $-$3.0E$+$1 & $+$3.0E$+$1 \\
Step & 0 & (0,\ 0,\ \ldots,\ 0) & $-$1.0E$+$2 & $+$1.0E$+$2 \\
Quartic & 0 & (0,\ 0,\ \ldots,\ 0) & $-$1.28 & $+$1.28 \\\midrule
Schwefel 2.26 & $-$1.257E$+$4 & (421,\ 421,\ \ldots,\ 421) & $-$5.0E$+$2 & $+$5.0E$+$2 \\
Rastrigin & 0 & (0,\ 0,\ \ldots,\ 0) & $-$5.12 & $+$5.12 \\
Ackley & 0 & (0,\ 0,\ \ldots,\ 0) & $-$3.2E$+$1 & $+$3.2E$+$1 \\
Griewank & 0 & (0,\ 0,\ \ldots,\ 0) & $-$6.0E$+$2 & $+$6.0E$+$2 \\
Penalized 1 & 0 & (1,\ 1,\ \ldots,\ 1) & $-$5.0E$+$1 & $+$5.0E$+$1 \\
Penalized 2 & 0 & (1,\ 1,\ \ldots,\ 1) & $-$5.0E$+$1 & $+$5.0E$+$1 \\\bottomrule
\end{tabular}
\end{table*}

\begin{sidewaystable}[htb]
\centering
\caption{Results from experiments with different values of the constant $\beta$}
\label{tab:normalized}
\begin{tabular}{ c c c c c c c c}\toprule
{\bf Function} & $\bm{\beta=0.125}$ & $\bm{\beta=0.25}$ & $\bm{\beta=0.5}$ & $\bm{\beta=1}$ & $\bm{\beta=2}$ & $\bm{\beta=4}$ & $\bm{\beta=8}$ \\\midrule
Sphere & 1.3599E$-$8 & {\it 5.3389E$-$18} & 2.107E$-$17 & 8.1714E$-$17 & 3.3475E$-$16 & 1.3119E$-$15 & 5.5299E$-$15 \\
Schwefel 2.22 & 7.3137E$-$1 & {\it 2.1712E$-$9} & 2.2372E$-$9 & 4.6915E$-$9 & 9.339E$-$9 & 1.9557E$-$8 & 3.5713E$-$8 \\
Schwefel 1.2 & 4.7587E$+$2 & 2.9554E$+$2 & 2.3724E$+$2 & {\it 2.1971E$+$2} & 4.0093E$+$2 & 6.3198E$+$2 & 6.2781E$+$2 \\
Schwefel 2.21 & 7.551 & 1.2665 & {\it 3.1007E$-$9} & 6.2955E$-$9 & 1.2524E$-$8 & 2.4962E$-$8 & 5.0208E$-$8 \\
Rosenbrock & 3.5013E$+$2 & 3.600E$+$1 & 2.6602E$+$1 & 2.6664E$+$1 & 2.7896E$+$1 & 2.6982E$+$1 & 2.5738E$+$1 \\
Step & 2.7773E$+$2 & 6.6667E$-$2 & {\it 0} & {\it 0} & {\it 0} & {\it 0} & {\it 0} \\
Quartic & 8.8352E$-$2 & 6.3546E$-$2 & 4.3361E$-$2 & 1.7258E$-$2 & {\it 1.5267E$-$2} & 1.7875E$-$2 & 1.7551E$-$2 \\\midrule
Schwefel 2.26 & $-$2.5884E$+$3 & $-$2.8147E$+$3 & $-$2.7307E$+$3 & $-$2.4462E$+$3 & $-$2.6851E$+$3 & $-$3.2117E$+$3 & {\it $-$3.268E$+$3} \\
Rastrigin & {\it 6.4424} & 6.5291 & 6.6644 & 8.1304 & 1.1634E$+$1 & 1.294E$+$1 & 1.3187E$+$1 \\
Ackley & 7.1229 & 2.1282E$-$8 & {\it 1.1419E$-$9} & 2.2832E$-$9 & 4.4431E$-$9 & 8.6187E$-$9 & 1.812E$-$8 \\
Griewank & 2.2833E$-$1 & 4.73E$-$14 & 6.4632E$-$16 & 7.2861E$-$16 & {\it 5.0886E$-$16} & 4.536E$-$15 & 6.4975E$-$15 \\
Penalized 1 & 2.2884 & 1.4927 & 5.4368E$-$7 & 1.3393E$-$16 & \textit{1.8198E$-$18} & 1.3062E$-$15 & 9.5627E$-$17 \\
Penalized 2 & 2.9307E$+$1 & 2.6607 & 4.0881E$-$12 & {\it 2.2543E$-$17} & 2.5413E$-$17 & 3.2931E$-$17 & 3.9909E$-$16 \\\bottomrule
\end{tabular}
\end{sidewaystable}

\clearpage

The results indicated that the $\beta$ value of 1.0 produces, in general, the best final solution, the fastest convergence, and the least number of occurrences of premature convergence. It is worth noticing that all values, from 1.0 to 8.0, produced good results in most functions, showing that GSA-NGC is quite insensitive to the value of $\beta$ in this range.

\section{Experiments and results}
\label{sec:results}

The experiments aim to compare the performance of GSA-NGC to that of GSA \cite{rashedi} and of two very recent versions of GSA --- GSAN \cite{khajooei} and FVGGSA \cite{bansal} --- in terms of quality of solution and speed of convergence. The performance in terms of computational time was not evaluated, for the iterative part of GSA-NGC is identical to that of GSA.

The GSA-NGC, with a $\beta$ value equal to 1.0, is compared to the GSA, GSAN and FVGGSA with the value of the initial gravitational constant $G_0$ equal to 100, the value used by Rashedi et al.\ \cite{rashedi}. The comparison is made on the thirteen reference functions described in \autoref{tab:reference1}. The result of the optimization of each function is given by the average of thirty executions of each algorithm.

Three series of experiments were performed. In the first, a square search space was used, in which each dimension $d_i$ of the space is 100 times smaller than dimension $d_i$ used by Rashedi et al.\ \cite{rashedi}. In the second, a square search space was used, in which each dimension $d_i$ of the space is 100 times larger than dimension $d_i$ used by Rashedi et al.\ \cite{rashedi}. Lastly, in the third series of experiments, a rectangular search space was used, in which each dimension $d_i$ is one order of magnitude smaller than the following one ($d_{i+1}$), according to the set of Equations~\ref{eq:minRect}

\begin{equation}
\begin{matrix}
L_i^-=L_0^-*10^{i-6}, \quad i=1,\ 2,\ ...,\ 11\\
L_i^+=L_0^+*10^{i-6}, \quad i=1,\ 2,\ ...,\ 11
\end{matrix}
\label{eq:minRect}
\end{equation}

\noindent where $L_i^-$ and $L_i^+$ are the minimum and maximum limits of dimension $d_i$ of the search space of the experiment, and $L_0^-$ and $L_0^+$ are the minimum and maximum limits of the search space used by Rashedi et al.\ \cite{rashedi}. The dimensions of the search space given by the set of Equations~\ref{eq:minRect} are shown in \autoref{tab:configuration3}.

\begin{table*}[htb]
\centering
\caption{Dimension of the search space and normalized gravitational constant ($G_0$) used on the third series of experiments}
\label{tab:configuration3}
\begin{tabular}{c c c c c c c}\toprule
\multirow{2}{*}{\bf Function} & 
\multicolumn{5}{c}{\bf Dimension of the search space} & 
\multirow{2}{*}{$\bm{G_0}$} \\ \cmidrule{2-6} &
\multicolumn{1}{c}{\bf 1} & 
\multicolumn{1}{c}{\bf 2} & 
\multicolumn{1}{c}{\bf 3} & 
\multicolumn{1}{c}{\bf \ldots} & 
\multicolumn{1}{c}{\bf 11} & \\\midrule
Sphere & $\pm$1E$-$3 & $\pm$1E$-$2 & $\pm$1E$-$1 & \ldots & $\pm$1E$+$7 & 2.0202E$+$6 \\
Schwefel 2.22 & $\pm$1E$-$4 & $\pm$1E$-$3 & $\pm$1E$-$2 & \ldots & $\pm$1E$+$6 & 2.0202E$+$5 \\
Schwefel 1.2 & $\pm$1E$-$3 & $\pm$1E$-$2 & $\pm$1E$-$1 & \ldots & $\pm$1E$+$7 & 2.0202E$+$6 \\
Schwefel 2.21 & $\pm$1E$-$3 & $\pm$1E$-$2 & $\pm$1E$-$1 & \ldots & $\pm$1E$+$7 & 2.0202E$+$6 \\
Rosenbrock & 1$\pm$3E$-$4 & 1$\pm$3E$-$3 & 1$\pm$3E$-$2 & \ldots & 1$\pm$3E$+$6 & 6.0606E$+$5 \\
Step & $\pm$1E$-$3 & $\pm$1E$-$2 & $\pm$1E$-$1 & \ldots & $\pm$1E$+$7 & 2.0202E$+$6 \\
Quartic & $\pm$1.28E$-$5 & $\pm$1.28E$-$4 & $\pm$1.28E$-$3 & \ldots & $\pm$1.28E$+$5 & 2.5859E$+$4 \\\midrule
Schwefel 2.26 & 421$\pm$5E$-$3 & 421$\pm$5E$-$2 & 421$\pm$5E$-$1 & \ldots & 421$\pm$5E$+$7 & 1.0101E$+$7 \\
Rastrigin & $\pm$5.12E$-$5 & $\pm$5.12E$-$4 & $\pm$5.12E$-$3 & \ldots & $\pm$5.12E$+$5 & 1.0343E$+$5 \\
Ackley & $\pm$3.2E$-$4 & $\pm$3.2E$-$3 & $\pm$3.2E$-$2 & \ldots & $\pm$3.2E$+$6 & 6.4646E$+$5 \\
Griewank & $\pm$6E$-$3 & $\pm$6E$-$2 & $\pm$6E$-$1 & \ldots & $\pm$6E$+$7 & 1.2121E$+$7 \\
Penalized 1 & 1$\pm$5E$-$4 & 1$\pm$5E$-$3 & 1$\pm$5E$-$2 & \ldots & 1$\pm$5E$+$6 & 1.0101E$+$6 \\
Penalized 2 & 1$\pm$5E$-$4 & 1$\pm$5E$-$3 & 1$\pm$5E$-$2 & \ldots & 1$\pm$5E$+$6 & 1.0101E$+$6 \\\bottomrule
\end{tabular}
\end{table*}

The two common parameters --- number of particles $P$ and number of iterations $T$ ---, shared by all algorithms, were assigned the values $50$ and $1000$, respectively. The third common parameter --- number of dimensions $D$ --- was assigned the value $30$ in the first and second series of experiments, and the value $11$ in the third series. The fourth common parameter --- $\alpha$ --- was assigned the value $20$ in GSA, GSAN and GSA-NGC, and the value $10$ in FVGGSA. The two specific parameters of GSAN --- $C_0$ and $\beta$ --- were assigned the values $0.1$ and $20$, respectively. The specific parameter of FVGGSA --- scaling constant $Z$ --- was assigned, for each phase, the values shown in \autoref{tab:FVGGSAP}. All these parameter values are identical to those used by the authors, as reported in the literature \cite{rashedi,khajooei,bansal}.

\begin{table*}[htb]
\centering
\caption{Values of the scaling constant $Z$ for each phase of FVGGSA}
\label{tab:FVGGSAP}
\begin{tabular}{c c c c c c}\toprule
Iterations & 1--200 & 201--400 & 401--600 & 601--800 & 801--1000 \\\midrule
Z & 100 & 0.5 & 0.3 & 0.2 & 0.01 \\\bottomrule
\end{tabular}
\end{table*}

\begin{figure*}[htb]
\includegraphics[width=0.5\linewidth]{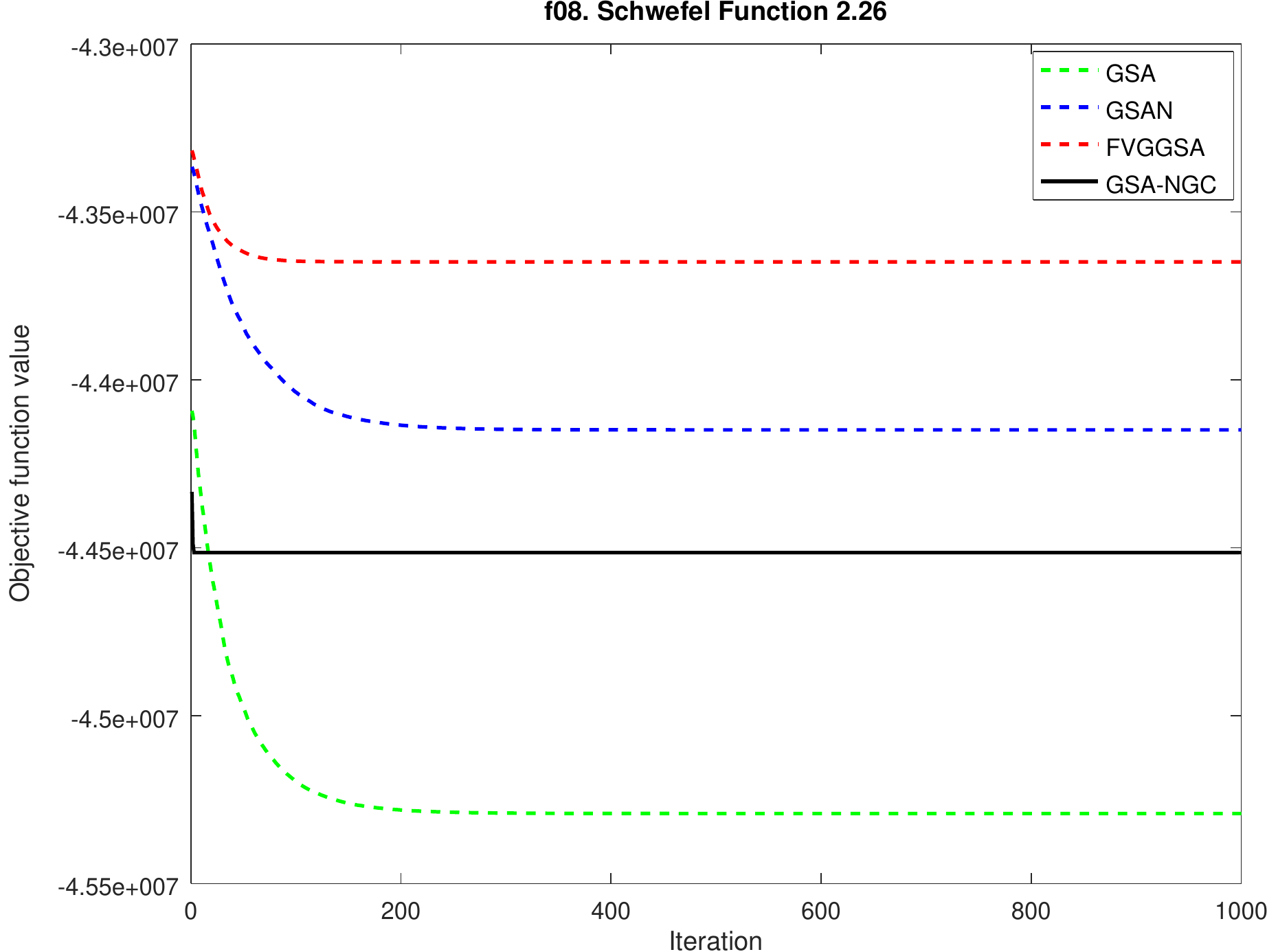}
\includegraphics[width=0.5\linewidth]{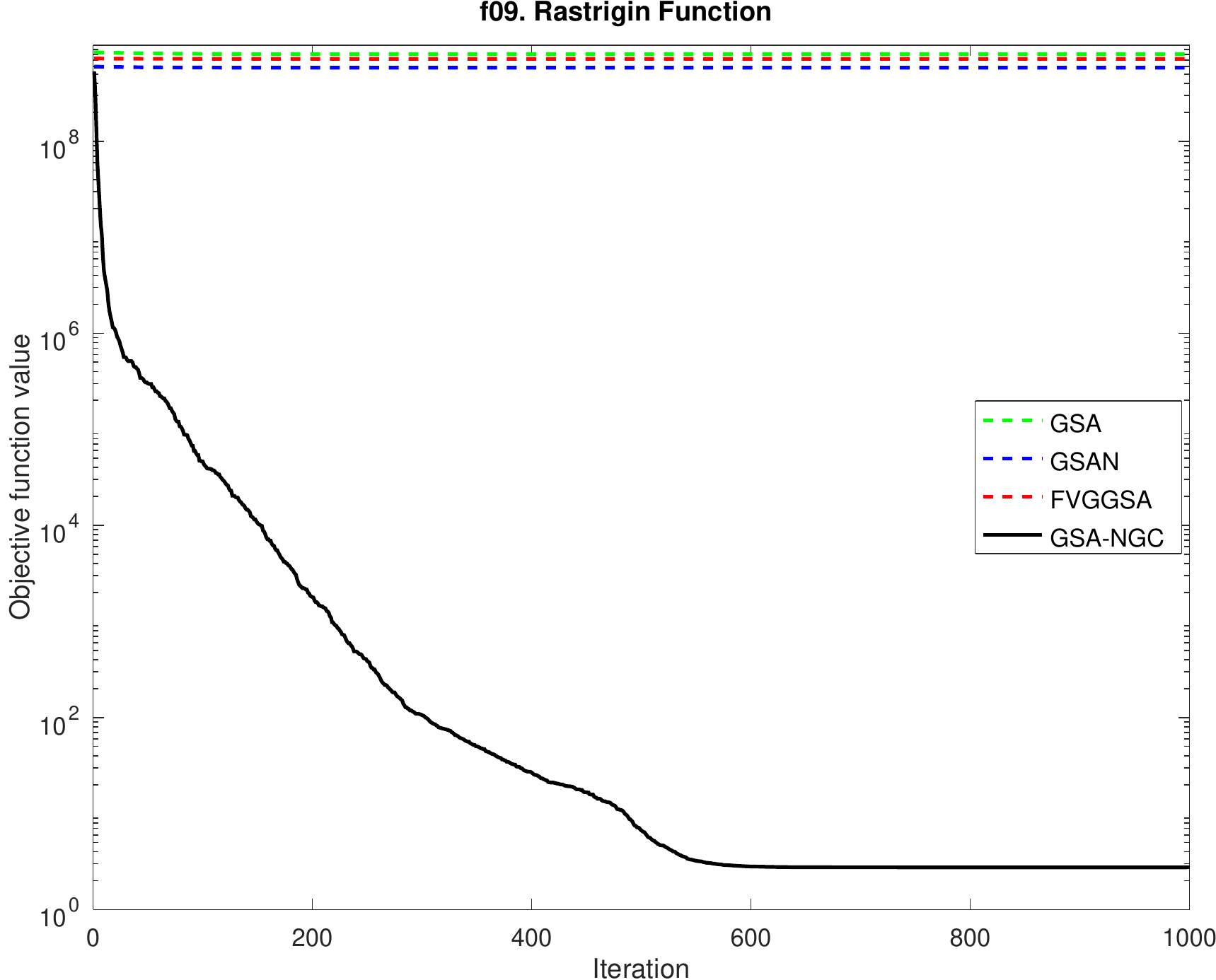}
\includegraphics[width=0.5\linewidth]{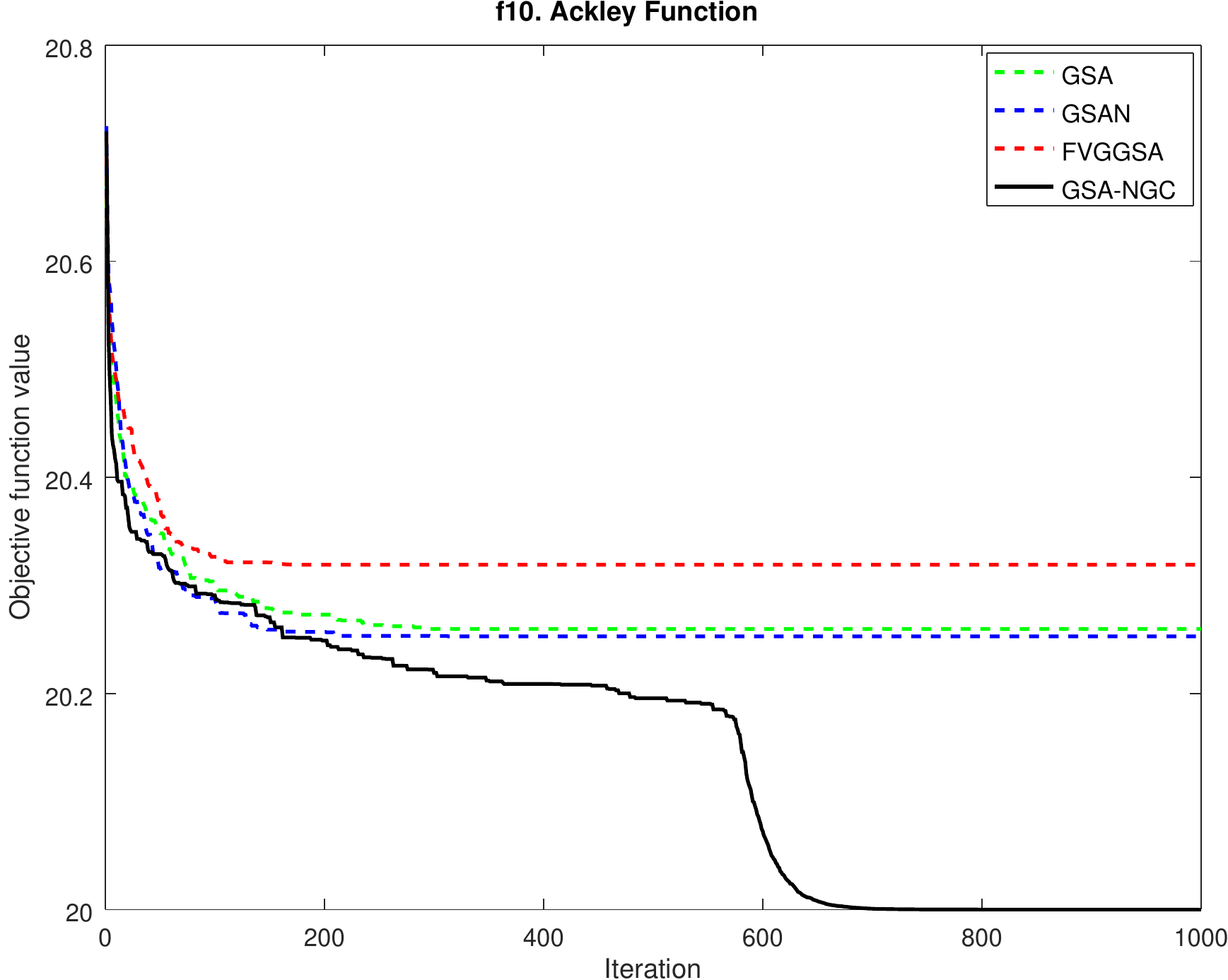}
\includegraphics[width=0.5\linewidth]{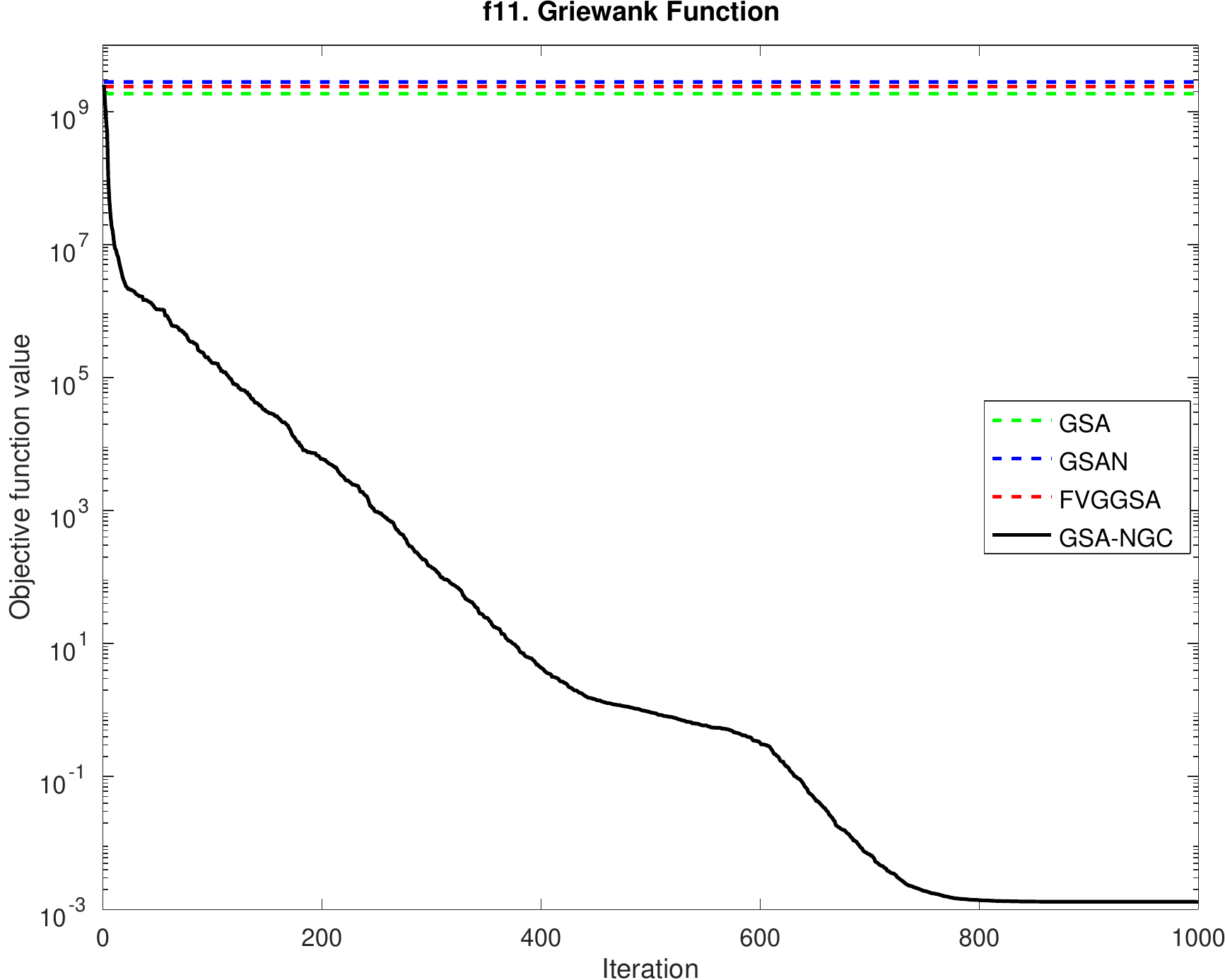}
\includegraphics[width=0.5\linewidth]{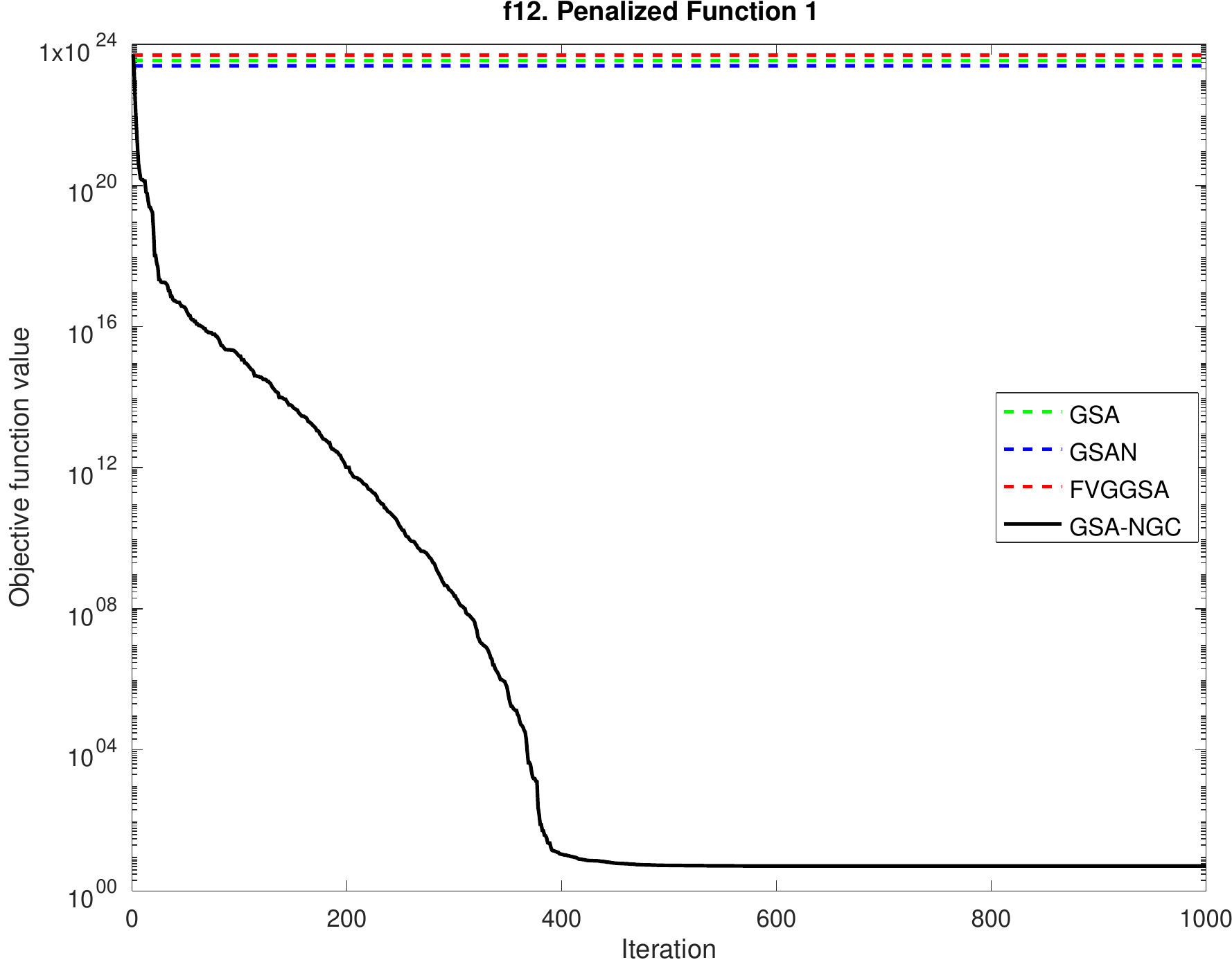}
\includegraphics[width=0.5\linewidth]{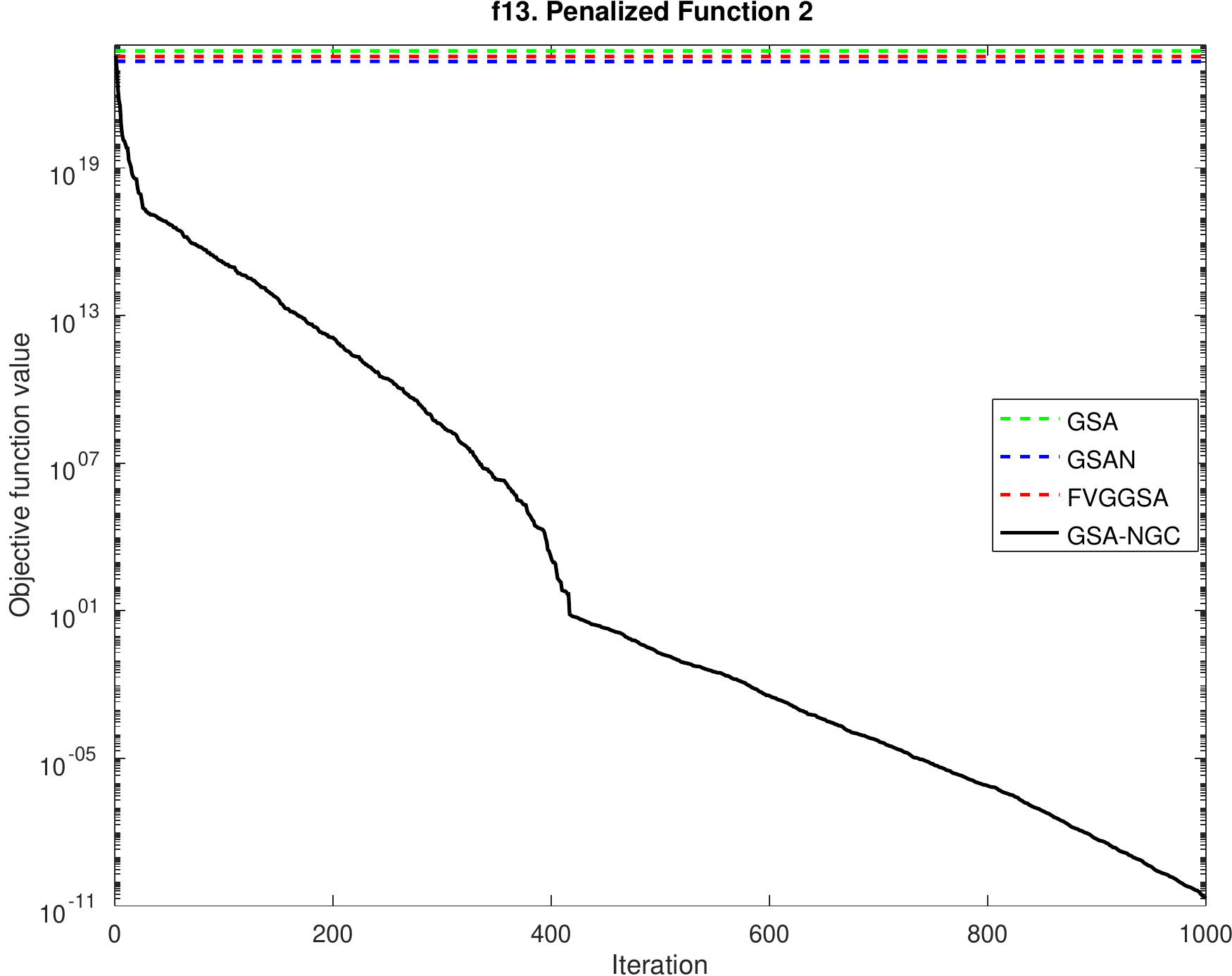}
\caption{Results from the third series of experiments, with the six multimodal functions}
\label{fig:experiment6}
\end{figure*}

\autoref{tab:experiment1}, \autoref{tab:experiment2} and \autoref{tab:experiment3} present, respectively, the results of the first, second and third series of experiments with the thirteen functions. \autoref{fig:experiment6}, in its turn, plots the results with only the six multimodal functions on the third series of experiments.

\begin{sidewaystable}[htb]
\centering
\caption{Results from the first series of experiments --- $\gamma$: confidence interval}
\label{tab:experiment1}
\begin{tabular}{c c c c c c c c c}\toprule
{\bf Function} & {\bf GSA} & {\bf GSA} $\bm{\gamma}$ & {\bf GSAN} & {\bf GSAN} $\bm{\gamma}$ &
    {\bf FVGGSA} & {\bf FVGGSA} $\bm{\gamma}$ & {\bf GSA-NGC} & {\bf GSA-NGC} $\bm{\gamma}$ \\\midrule
Sphere & 2.30E$-$17 & $\pm$1.82E$-$18 & 2.97E$-$17 & $\pm$7.25E$-$18 &
    1.22E$-$17 & $\pm$1.14E$-$18 & {\it 8.54E$-$21} & $\pm$7.22E$-$22 \\
Schwefel 2.22 & 2.30E$-$08 & $\pm$1.17E$-$09 & 2.81E$-$08 & $\pm$2.54E$-$09 &
    1.77E$-$08 & $\pm$7.29E$-$10 & {\it 4.70E$-$11} & $\pm$3.52E$-$12 \\
Schwefel 1.2 & 5.83E$-$02 & $\pm$7.60E$-$03 & {\it 6.14E$-$08} & $\pm$9.26E$-$08 &
    2.16E$-$01 & $\pm$2.41E$-$02 & 2.66E$-$02 & $\pm$2.86E$-$03 \\
Schwefel 2.21 & 3.29E$-$09 & $\pm$3.34E$-$10 & 2.21E$-$08 & $\pm$4.91E$-$09 &
    2.53E$-$03 & $\pm$8.77E$-$04 & {\it 6.70E$-$11} & $\pm$4.19E$-$12 \\
Rosenbrock & 1.68E$-$03 & $\pm$6.60E$-$04 & 1.68E$-$03 & $\pm$6.60E$-$04 &
    2.62E$-$02 & $\pm$9.05E$-$03 & {\it 2.99E$-$04} & $\pm$1.13E$-$04 \\
Step & {\it 0} & 0 & {\it 0} & 0 & {\it 0} & 0 & {\it 0} & 0 \\
Quartic & 2.75E$-$05 & $\pm$9.03E$-$06 & 1.96E$-$05 & $\pm$6.76E-06 &
    2.81E$-$05 & $\pm$8.50E$-$06 & 1.90E$-$05 & $\pm$8.23E$-$06 \\\midrule
Schwefel 2.26 & $-$1.26E$+$04 & $\pm$1.99E$-$12 & $-$1.26E$+$04 & $\pm$1.99E$-$12 &
    $-$1.26E$+$04 & $\pm$1.99E$-$12 & $-$1.26E$+$04 & $\pm$1.99E$-$12 \\
Rastrigin & 7.11E$-$15 & $\pm$9.59E$-$16 & 1.14E$-$14 & $\pm$2.59E$-$15 &
    2.31E$-$15 & $\pm$7.11E$-$16 & {\it 0} & 0 \\
Ackley & 3.61E$-$09 & $\pm$2.09E$-$10 & 4.00E$-$09 & $\pm$3.49E$-$10 &
    2.80E$-$09 & $\pm$1.22E$-$10 & {\it 2.28E$-$11} & $\pm$1.32E$-$12 \\
Griewank & {\it 0} & 0 & 9.86E$-$04 & $\pm$9.15E$-$04 &
    2.90E$-$03 & $\pm$1.36E$-$03 & {\it 0} & 0 \\
Penalized 1 & {\it 5.54} & $\pm$2.86E$-$02 & 5.80 & $\pm$4.86E$-$02 &
    6.11 & $\pm$5.28E$-$02 & 6.76 & $\pm$7.13E$-$02 \\
Penalized 2 & 2.11E$-$18 & $\pm$1.76E$-$19 & 3.58E$-$18 & $\pm$1.57E$-$18 &
    9.32E$-$05 & $\pm$5.61E$-$05 & {\it 2.06E$-$22} & $\pm$1.78E$-$23 \\\bottomrule
\end{tabular}
\end{sidewaystable}

\begin{sidewaystable}[htb]
\centering
\caption{Results from the second series of experiments --- $\gamma$: confidence interval}
\label{tab:experiment2}
\begin{tabular}{c c c c c c c c c}\toprule
{\bf Function} & {\bf GSA} & {\bf GSA} $\bm{\gamma}$ & {\bf GSAN} & {\bf GSAN} $\bm{\gamma}$ &
    {\bf FVGGSA} & {\bf FVGGSA} $\bm{\gamma}$ & {\bf GSA-NGC} & {\bf GSA-NGC} $\bm{\gamma}$ \\\midrule
Sphere & 5.13E$+$08 & $\pm$1.87E$+$07 & 5.83E$+$08 & $\pm$2.12E$+$07 &
    6.33E$+$08 & $\pm$2.39E$+$07 & {\it 9.23E$-$13} & $\pm$8.26E-14 \\
Schwefel 2.22 & 1.87E$+$03 & $\pm$9.01E$+$01 & 8.40E$+$66 & $\pm$7.99E$+$66 &
    5.56E$+$62 & $\pm$6.60E$+$62 & {\it 8.76E$+$02} & $\pm$1.06E$+$02 \\
Schwefel 1.2 & 1.11E$+$09 & $\pm$1.02E$+$08 & 1.11E$+$09 & $\pm$1.29E$+$08 &
    1.36E$+$09 & $\pm$1.64E$+$08 & {\it 2.73E$+$06} & $\pm$3.95E$+$05 \\
Schwefel 2.21 & 7.70E$+$03 & $\pm$1.02E$+$02 & 8.00E$+$03 & $\pm$1.33E$+$02 &
    8.41E$+$03 & $\pm$1.32E$+$02 & {\it 5.98E$-$07} & $\pm$3.85E$-$08 \\
Rosenbrock & 3.24E$+$15 & $\pm$2.47E$+$14 & 1.27E$+$16 & $\pm$1.10E$+$15 &
    1.88E$+$16 & $\pm$1.17E$+$15 & {\it 5.95E$+$03} & $\pm$3.76E$+$03 \\
Step & 4.93E$+$08 & $\pm$2.19E$+$07 & 6.11E$+$08 & $\pm$1.94E$+$07 &
    6.03E$+$08 & $\pm$2.77E$+$07 & {\it 0} & 0 \\
Quartic & 4.41E$-$02 & $\pm$1.07E$-$02 & 9.48E$-$02 & $\pm$1.01E$-$02 &
    1.88E$+$05 & $\pm$8.26E$+$04 & {\it 2.74E$-$02} & $\pm$4.47E$-$03 \\\midrule
Schwefel 2.26 & $-$2.98E$+$05 & $\pm$1.58E$+$04 & $-$2.83E$+$05 & $\pm$1.95E$+$04 &
    $-$2.57E$+$05 & $\pm$1.65E$+$04 & $-$2.64E$+$05 & $\pm$1.82E$+$04 \\
Rastrigin & 6.05E$+$03 & $\pm$1.14E$+$03 & 1.46E$+$05 & $\pm$1.77E$+$04 &
    6.54E$+$05 & $\pm$3.18E$+$04 & {\it 1.83E$+$01} & $\pm$1.82 \\
Ackley & 2.11E$+$01 & $\pm$1.62E$-$02 & 2.11E$+$01 & $\pm$2.33E$-$02 &
    2.11E$+$01 & $\pm$2.96E$-$02 & {\it 2.00E$+$01} & $\pm$1.50E$-$14 \\
Griewank & 5.58E$+$06 & $\pm$2.16E$+$05 & 5.77E$+$06 & $\pm$2.38E$+$05 &
    5.83E$+$06 & $\pm$1.90E$+$05 & {\it 5.75E$-$04} & $\pm$7.92E$-$04 \\
Penalized 1 & 6.37E$+$16 & $\pm$4.41E$+$15 & 1.29E$+$17 & $\pm$7.84E$+$15 &
    1.59E$+$17 & $\pm$9.69E$+$15 & {\it 3.46E$-$03} & $\pm$6.77E$-$03 \\
Penalized 2 & 6.50E$+$16 & $\pm$4.29E$+$15 & 1.29E$+$17 & $\pm$8.66E$+$15 &
    1.58E$+$17 & $\pm$1.11E$+$16 & {\it 3.66E$-$04} & $\pm$7.18E$-$04 \\\bottomrule
\end{tabular}
\end{sidewaystable}

\begin{sidewaystable}[htb]
\centering
\caption{Results from the third series of experiments --- $\gamma$: confidence interval}
\label{tab:experiment3}
\begin{tabular}{c c c c c c c c c}\toprule
{\bf Function} & {\bf GSA} & {\bf GSA} $\bm{\gamma}$ & {\bf GSAN} & {\bf GSAN} $\bm{\gamma}$ &
    {\bf FVGGSA} & {\bf FVGGSA} $\bm{\gamma}$ & {\bf GSA-NGC} & {\bf GSA-NGC} $\bm{\gamma}$ \\\midrule
Sphere & 3.31E$+$11 & $\pm$1.33E$+$11 & 2.77E$+$11 & $\pm$8.20E$+$10 &
    1.77E$+$11 & $\pm$7.83E$+$10 & {\it 7.74E$-$10} & $\pm$8.94E$-$11 \\
Schwefel 2.22 & 9.99E$+$04 & $\pm$2.92E$+$04 & 8.64E$+$04 & $\pm$2.45E$+$04 &
    1.08E$+$05 & $\pm$2.11E$+$04 & {\it 9.40E$-$06} & $\pm$6.10E$-$07 \\
Schwefel 1.2 & 3.40E$+$11 & $\pm$1.34E$+$11 & 3.08E$+$11 & $\pm$1.25E$+$11 &
    2.72E$+$11 & $\pm$1.12E$+$11 & {\it 2.20E$-$09} & $\pm$3.12E$-$10 \\
Schwefel 2.21 & 3.50E$+$05 & $\pm$5.89E$+$04 & 4.00E$+$05 & $\pm$7.95E$+$04 &
    3.96E$+$05 & $\pm$6.35E$+$04 & {\it 2.01E$-$05} & $\pm$1.68E$-$06 \\
Rosenbrock & 4.88E$+$18 & $\pm$2.57E$+$18 & 5.42E$+$18 & $\pm$4.01E$+$18 &
    1.16E$+$19 & $\pm$7.74E$+$18 & {\it 7.78E$+$12} & $\pm$5.00E$+$12 \\
Step & 2.18E$+$11 & $\pm$5.91E$+$10 & 2.03E$+$11 & $\pm$7.80E$+$10 &
    2.00E$+$11 & $\pm$6.21E$+$10 & {\it 0} & 0 \\
Quartic & 2.10E$+$16 & $\pm$1.65E$+$16 & 9.96E$+$15 & $\pm$6.38E$+$15 &
    8.96E$+$15 & $\pm$5.79E$+$15 & {\it 4.43E$-$03} & $\pm$7.09E$-$04 \\\midrule
Schwefel 2.26 & $-$4.53E$+$07 & $\pm$1.62E$+$06 & $-$4.41E$+$07 & $\pm$1.73E$+$06 &
    $-$4.36E$+$07 & $\pm$1.67E$+$06 & $-$4.45E$+$07 & $\pm$1.24E$+$06 \\
Rastrigin & 8.07E$+$08 & $\pm$2.90E$+$08 & 5.82E$+$08 & $\pm$2.24E$+$08 &
    7.21E$+$08 & $\pm$2.61E$+$08 & {\it 2.75} & $\pm$6.52E$-$01 \\
Ackley & 2.03E$+$01 & $\pm$2.70E$-$02 & 2.03E$+$01 & $\pm$2.83E$-$02 &
    2.03E$+$01 & $\pm$3.20E$-$02 & {\it 2.00E$+$01} & $\pm$6.28E$-$11 \\
Griewank & 1.87E$+$09 & $\pm$5.95E$+$08 & 2.79E$+$09 & $\pm$9.54E$+$08 &
    2.39E$+$09 & $\pm$8.11E$+$08 & {\it 1.31E$-$03} & $\pm$2.57E$-$03 \\
Penalized 1 & 3.44E$+$23 & $\pm$2.09E$+$23 & 2.45E$+$23 & $\pm$9.11E$+$22 &
    4.90E$+$23 & $\pm$2.54E$+$23 & {\it 5.11} & $\pm$9.82E$-$04 \\
Penalized 2 & 5.76E$+$23 & $\pm$3.38E$+$23 & 2.16E$+$23 & $\pm$1.09E$+$23 &
    3.39E$+$23 & $\pm$2.42E$+$23 & {\it 2.05E$-$11} & $\pm$3.13E$-$12 \\\bottomrule
\end{tabular}
\end{sidewaystable}

\clearpage

\clearpage

\autoref{tab:experiment1} shows that all algorithms produced statistically equivalent solutions, considering the confidence interval, on the functions Quartic and Schwefel 2.26. GSAN and FVGGSA reached the optimum value on just one function --- Step. GSA reached the optimum value on two functions --- Step and Griewank ---, and GSA-NGC on three functions --- Step, Rastrigin, and Griewank. GSAN and GSA produced better solutions than those produced by the other algorithms on the functions Schwefel 1.2 and Penalized 1, respectively. In its turn, the solutions of GSA-NGC were better than those of the other algorithms on the Rastrigin and on the remaining six functions.

Both \autoref{tab:experiment2} and \autoref{tab:experiment3} convey the same analysis of their results. They show that all algorithms produced statistically equivalent solutions, considering the confidence interval, on the function Schwefel 2.26. GSA-NGC was the unique algorithm that reached the optimum value on a function --- Step. Moreover, the solutions of GSA-NGC were either better or much better than those of the other algorithms on the Step and on the remaining eleven functions.

The results shown in Tables \ref{tab:experiment1}, \ref{tab:experiment2}, \ref{tab:experiment3}, and in \autoref{fig:experiment6} indicate that the final solution of the GSA-NGC is much better than those of the GSA, GSAN and FVGGSA in practically all functions, regardless of the search space size and shape. In some cases, the final solution of GSA-NGC was more than 1.0E$+$20 times better than those of the other three algorithms. In addition, in these functions, convergence of the GSA-NGC occurred in a number of iterations much smaller than those of the other three algorithms.

\section{Conclusion}
\label{sec:conclusion}

This paper proposes the Gravitational Search Algorithm with Normalized Gravitational Constant (GSA-NGC), which defines a new heuristic to determine the initial gravitational constant, based on the multiple dimensions of the search space of the application. An initial gravitational constant even more suitable than that of the GSA-NGC could be determined empirically for each application. However, the empirical determination of this constant would require dozens of trials and corrections, whereas the proposed heuristic requires a very low processing cost for its calculation.

The GSA-NGC is experimentally validated by minimizing thirteen well-known reference functions from the literature, modified for squared and rectangular spaces of dimensions with highly irregular sizes. The results indicate that the GSA-NGC produces a major improvement in the final solutions and a major reduction in the number of iterations and of premature convergences when compared to the GSA, GSAN and FVGGSA.

As future work, we propose the study of the possible dependencies between different parameters of GSA, the analysis of the GSA-NGC heuristic in other versions of the GSA and in other evolutionary algorithms, as well as the possibility of using an adaptable $\alpha$ value, in order to control the variation of the gravity.